\pdfoutput=1

\documentclass[11pt]{article}

\usepackage[final]{ACL2023}

\usepackage{times}
\usepackage{latexsym}

\usepackage[T1]{fontenc}

\usepackage[utf8]{inputenc}

\usepackage{microtype}

\usepackage{inconsolata}

\usepackage{amsmath}
\usepackage{tabularx}
\newcolumntype{L}[1]{>{\raggedright\arraybackslash}p{#1}}%
\newcolumntype{R}[1]{>{\raggedleft\arraybackslash}p{#1}}%
\newcolumntype{C}[1]{>{\centering\arraybackslash}p{#1}}%

\usepackage{subcaption}
\usepackage{float}
\usepackage{booktabs}
\usepackage[flushleft]{threeparttable}
\usepackage{multirow}
\usepackage{amsmath}
\usepackage{amssymb}
\usepackage[inline]{enumitem}
\setlist[description]{leftmargin=2ex,leftmargin=!,parsep=0pt,itemsep=1pt,topsep=1pt}  
\setlist[enumerate]{labelindent=0ex,labelwidth=3ex,leftmargin=!,parsep=0pt,itemsep=1pt,topsep=1pt}
\setlist[itemize]{labelindent=0ex,labelwidth=1.5ex,leftmargin=!,parsep=0pt,itemsep=1pt,topsep=1pt}
\usepackage{mathtools}

\usepackage{cleveref}
\crefname{section}{Section}{Sections}
\Crefname{section}{Section}{Sections}
\crefname{appendix}{Appendix}{Appendices}
\crefname{Appendix}{Appendix}{Appendices}
\crefname{subsection}{Section}{Sections}
\Crefname{subsection}{Section}{Sections}
\crefname{equation}{Equation}{Equations}
\Crefname{equation}{Equation}{Equations}
\Crefname{figure}{Figure}{Figures}
\crefname{figure}{Figure}{Figures}
\Crefname{table}{Table}{Tables}
\crefname{table}{Table}{Tables}
\crefname{enumi}{}{}
\crefname{algorithm}{Algorithm}{Algorithms}
\Crefname{algorithm}{Algorithm}{Algorithms}

\usepackage{autonum} 

\title{Hitachi at SemEval-2023 Task 3: Exploring Cross-lingual Multi-task Strategies for Genre and Framing Detection in Online News}

\author{Yuta Koreeda\thanks{\quad Equal contribution},~
        Ken-ichi Yokote\footnotemark[1],~
        Hiroaki Ozaki,\\
        \textbf{Atsuki Yamaguchi},~
        \textbf{Masaya Tsunokake} \and
        \textbf{Yasuhiro Sogawa} \\
        Research and Development Group, Hitachi, Ltd.\\
        Kokubunji, Tokyo, Japan\\
        \texttt{\{yuta.koreeda.pb, kenichi.yokote.fb, hiroaki.ozaki.yu,}\\
        \texttt{atsuki.yamaguchi.xn, masaya.tsunokake.qu, yasuhiro.sogawa.tp\}@hitachi.com}
        }

\begin{document}

\maketitle
\begin{abstract}
This paper explains the participation of team \emph{Hitachi} to SemEval-2023 Task 3 ``\emph{Detecting the genre, the framing, and the persuasion techniques in online news in a multi-lingual setup}.''
Based on the multilingual, multi-task nature of the task and the low-resource setting, we investigated different cross-lingual and multi-task strategies for training the pretrained language models.
Through extensive experiments, we found that \begin{enumerate*}[label=(\alph*)]
    \item cross-lingual/multi-task training, and
    \item collecting an external balanced dataset
\end{enumerate*}, can benefit the genre and framing detection.
We constructed ensemble models from the results and achieved the highest macro-averaged F1 scores in Italian and Russian genre categorization subtasks.

\end{abstract}

\section{Introduction}\label{sec:introduction}

As we pay more and more attention to the socially influencing problems like COVID-19 and the Russo-Ukrainian war, there has been an increasing concern about \emph{infodemic} of false and misleading information \cite{semeval2023task3}.
In particular, cross-lingual understanding of such information is becoming more important due to polarization of political stances, economical decoupling and echo chamber effect in social media.
To that end, \citet{semeval2023task3} put together SemEval-2023 Task 3 ``\emph{Detecting the genre, the framing, and the persuasion techniques in online news in a multi-lingual setup}.''
The shared task aims to analyze several aspects of what makes a text persuasive and to foster development of building blocks for multilingual media analysis.

Creating annotated data for media analysis is time consuming thus we cannot assume that we can obtain training data of enough quality and quantity.
To tackle the problem, we investigated and compared strategies for multilingual media analysis under a low-resource setting.
Through extensive experiments, we found that \begin{enumerate*}[label=(\alph*)]
    \item cross-lingual/multi-task training, and
    \item collecting an external balanced dataset
\end{enumerate*}, can benefit the genre and framing detection.
We constructed ensemble models from the results and participated in genre categorization (subtask 1) and framing detection (subtask 2) in six languages, where we achieved the highest macro-averaged F1 scores in Italian and Russian subtask 1.

\section{Task Definition and our Strategy}\label{sec:task}

SemEval-2023 Task 3 aims to analyze several aspects of what makes a text persuasive.
It offers three subtasks on news articles in six languages: German (de), English (en), French (fr), Italian (it), Polish (pl) and Russian (ru). There are three additional languages (Georgian, Greek and Spanish) without training datasets (i.e., participants need to perform zero-shot language transfer).

\begin{description}
    \item[Subtask 1: News genre categorization] Given a news article, a system has to determine whether it is an opinion piece, it aims at objective news reporting, or it is a satire piece. This is multi-class document classification and the official evaluation measure is macro average F1 score (macro-F1) over the three classes.
    \item[Subtask 2: Framing detection] Given a news article, a system has to identify what key aspects (frames) are highlighted the rhetoric from 14 frames (see \cite{card-etal-2015-media} for the taxonomy and definitions). This is multi-label document classification and the official evaluation measure is micro average F1 score (micro-F1) over the 14 frames.
    \item[Subtask 3: Persuasion techniques detection] Given a news article, a system has to identify the persuasion techniques in each paragraph from 23 persuasion techniques. This is multi-label paragraph classification.
\end{description}

The target articles are those identified to be potentially spreading mis-/disinformation and are collected from 2020 to mid-2022.
They revolve around widely discussed topics such as COVID-19, migration, the build-up leading to the Russo-Ukrainian war, and some country-specific local events such as elections.

We observed that the numbers of articles are limited for the relative large label space and there exist considerable overlaps of articles between subtask 1 and 2 (see \cref{sec:appendix-task}).
Hence, we decided to investigate if models trained on multiple languages or another subtask can benefit the target task in this low resource setting (\cref{sec:framework}).
Since subtask 1 and 2 conveniently share the task format, we opted to participate in subtask 1 and 2 in all the six languages.

We also noticed that the English training dataset exhibits significantly different label distribution to other languages and it is unbalanced.
Hence, we decided to collect additional external dataset for English subtask 1 in a wish to improve task performance in English and to help with other languages through cross-lingual training (\cref{sec:dataset}).

\section{External Data for English Genre Categorization}\label{sec:dataset}

In a preliminary analysis of the English subtask 1 dataset, we found that label distribution is quite unbalanced and it is different in the training and the development data.
Therefore, we did not make any assumption about the distribution of the test data and decided to increase the number of rare labels in order to create a new, balanced dataset for English genre categorization.
First, we undersampled articles from the training dataset for subtask 1 such that the numbers of articles for each label are equal, i.e., ten articles for each label.

\begin{table}[t!]
    \centering
    \fontsize{8pt}{8pt}\selectfont
    \setlength{\tabcolsep}{4pt}
    \renewcommand{\arraystretch}{1.2}
    \renewcommand{\aboverulesep}{0.3ex}  
    \renewcommand{\belowrulesep}{0.5ex}  
    \begin{tabularx}{\linewidth}{lX}
      \toprule
        Label & News media \\\midrule
        Satire & The Onion, Huffington Post Satire, Borowitz Report, The Beaverton, Satire Wire, and Faking News \\
        Reporting & Wall Street Journal, The Economist, BBC, NPR, ABC, CBS, USA Today, The Guardian, NBC, The Washington Post \\
        Opinion & Ending The Fed, True Pundit, abcnews.com.co, DC Gazette, Liberty Writers News, Before its News, InfoWars, Real News Right Now\\
        \bottomrule
    \end{tabularx}
    \caption{The list of news media that we idependently collected the data from}\label{tab:media-list}
\end{table}

\begin{table}[t!]
    \centering
    \begin{threeparttable}
        \centering
        \fontsize{8pt}{10pt}\selectfont
        \setlength{\tabcolsep}{3.2pt}
        \renewcommand{\arraystretch}{.6}
        \renewcommand{\aboverulesep}{0.3ex}  
        \renewcommand{\belowrulesep}{0.6ex}  
        \begin{tabular}{lrrrrrrrrrr}
          \toprule
           & \multicolumn{3}{c}{Satire} & \multicolumn{3}{c}{Reporting} & \multicolumn{3}{c}{Opinion} \\\cmidrule(lrr){2-4}\cmidrule(lrr){5-7}\cmidrule(lrr){8-10}
            Dataset name   & \multicolumn{1}{c}{O} & \multicolumn{1}{c}{E} & \multicolumn{1}{c}{C} & \multicolumn{1}{c}{O} & \multicolumn{1}{c}{E} & \multicolumn{1}{c}{C} & \multicolumn{1}{c}{O} & \multicolumn{1}{c}{E} & \multicolumn{1}{c}{C} & \multicolumn{1}{c}{Total} \\\midrule
            Original & 10 & 0 & 0 & 41 & 0 & 0 & 382 & 0 & 0 & 433 \\
            Augmented (small) & 10 & 75 &  0 & 10 & 75 &  0 & 10 & 75 &  0 & 255 \\
            Augmented (large) & 10 & 75 & 31 & 41 & 75 &  0 & 41 & 75 &  0 & 348 \\
            \bottomrule
        \end{tabular}
        \makeatletter\def\TPT@hsize{}\makeatletter
        \begin{tablenotes}[para,flushleft]
            \raggedright
            \fontsize{7pt}{7pt}\selectfont
            O: Official dataset, E: External, existing datasets, C: Collected by us
        \end{tablenotes}
    \end{threeparttable}
    \caption{Number of articles from different sources in the original and our augmented datasets}\label{tab:numbers-augmented}
\end{table}

We referred to a survey on fake news detection datasets \cite{d2021fake} and checked a total of 27 datasets to see if they can be converted to subtask 1 dataset format using the following criteria:
\begin{description}
    \item[Label similarity]
We checked whether the labels defined by an external dataset are close to subtask 1. For example, we focused on whether they used identical label names, such as ``satire''.
    \item[Text similarity]
We checked if the text type of a dataset is similar to subtask 1, such as whether they use news articles.
    \item[Task similarity]
We checked whether the task setting employed by a dataset is a method of classifying them into different classes rather than, for example, scoring them with a scale of 1 to 5.
\end{description}

After these checks, we adopted the Random Political News Data \cite{horne2017just} which contains 75 articles for each of three labels.
We added the total of 225 articles to the sampled 30 original articles and constructed the \emph{Augmented (small)} dataset which contains 255 articles in total.

Since \citet{horne2017just} disclose the news media from which the data was collected, we independently collected around 1,000 additional articles from the sources shown in \cref{tab:media-list}.
However, we found in a preliminary experiment that overloading the dataset with external sources did not improve the performance.
Hence we sampled 31 satire articles from the collected data and sampled more articles from the original dataset.
This resulted in \emph{Augmented (large)} with 348 articles altogether.
The final compositions of the augmented datasets are summarized in \cref{tab:numbers-augmented}.

Since \citet{horne2017just} considered English articles only, we were only able to obtain external data for English subtask 1.
Nevertheless, the augmented data might be able to benefit non-English and subtask 2 datasets through pretraining on the augmented English dataset.

\begin{figure*}[t]
    \centering
    \includegraphics{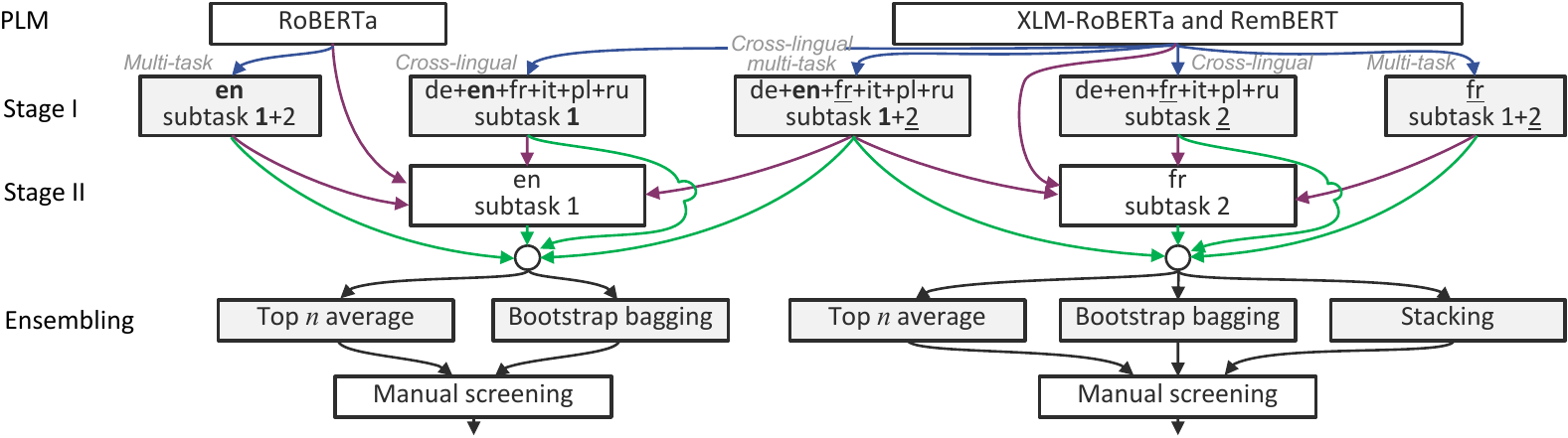}
    \caption{Series of random hyperparameter searches for exploring cross-lingual multi-task strategies. It only shows English subtask 1 and French subtask 2 but we did the same for all other languages in subtask 1 and 2.}\label{fig:training}
\end{figure*}

\section{Cross-lingual Multi-task Transformers}\label{sec:system}

We utilized pretrained language models (PLMs) in a simple sequence classification setup \cite{devlin_bert_2019}.
We employed XLM-RoBERTa large\footnote{\url{https://huggingface.co/}\{\texttt{xlm-roberta-large}, \texttt{rembert}, \texttt{roberta-large}\}} \cite{conneau-etal-2020-unsupervised} and RemBERT\footnotemark[\thefootnote] \cite{chung2021rethinking}.
For English, we also utilized RoBERTa large\footnotemark[\thefootnote] \cite{roberta} for single-language multi-task training.
In order to allow multi-task training, we added a classifier head for each subtask\footnote{dropout$\rightarrow$linear$\rightarrow$tanh$\rightarrow$dropout$\rightarrow$linear for (XLM-) RoBERTa, and dropout$\rightarrow$linear for RemBERT} on top of each model's \texttt{[CLS]} token followed by softmax for subtask 1 and sigmoid for subtask 2.
Hence, each model shares most of the parameters for the two subtasks.
We used Transformers library \cite{wolf-etal-2020-transformers} for the implementations.

In multi-task training we simply took sum of losses from two tasks.
Since there exist articles that only have either of subtask 1 or 2 labels, we ignore predictions for missing labels from loss calculation.
For cross-lingual training, we concatenate all articles.
All parameters are shared for the same subtask in different languages.
For preprocessing, we concatenated all sentences from each article and tokenized them using the default tokenizer for each PLM.
We truncated articles whose tokens do not fit onto each model's maximum context size.

\section{Exploring Cross-lingual Multi-task Strategies}\label{sec:framework}

It is empirically known that further fine-tuning a model trained in a multi-task or cross-lingual setting on the target downstream task/language improves its performance \cite{koreeda-etal-2019-hitachi}.
Also, models tend to require different hyperparameters for different training paradigms or languages.
Hence, we decided to explore different multi-task and cross-lingual strategies through a series of random hyperparameter searches (\cref{fig:training}).

First, we ran a random hyperparameter search in cross-lingual and/or multi-task settings (Stage I).
Regarding the resulting Stage I models as an additional hyperparameter, we ran another random hyperparameter search to optimize the choice of the pretraining paradigm along with other hyperparameters (Stage II).
Finally, we construct an ensemble for each language-subtask pair from all models in Stage I and II using their performance in the development dataset.

Unlike more sophisticated hyperparameter search methods, this approach has an advantage that we can compare and evaluate different training paradigms post hoc.

The choice of the subtask 1 English datasets (\cref{sec:dataset}) is also incorporated as an additional hyperparameter.
The hyperparameter search spaces are shown in \cref{sec:appendix-hyperparameters}.

We used the official development dataset for each language-subtask pair in order to calculate and compare the performance of each model.

\subsection{Stage I Training}\label{sec:framework-stage1}

In Stage I, we fine-tuned PLM in three settings.
\begin{enumerate}[label=(\arabic*)]
    \item Multi-task (30 hyperparameter sets for each language $=$ 180 models)
    \item Cross-lingual (50 hyperparameter sets for each subtask $=$ 100 models)
    \item Cross-lingual multi-task (50 hyperparameter sets $=$ 50 models)
\end{enumerate}
Hence, we trained 330 models in Stage I.

\subsection{Stage II Training}\label{sec:framework-stage2}

Stage I results in three groups of models that have been trained on each language-subtask pair.
For example, ``en subtask 1'' in \cref{fig:training} has incoming arrows from \begin{enumerate*}[label=(\arabic*)]
    \item multi-task (``en subtask 1+2''),
    \item cross-lingual (``de+en+fr+it+pl+ru subtask 1''), and
    \item cross-lingual multi-task (``de+en+fr+it+pl+ru subtask 1+2'')
\end{enumerate*}.
We also utilize vanilla PLMs for Stage II training (see the arrow from RoBERTa).

\begin{table*}[t]
    \centering
    \fontsize{8pt}{10pt}\selectfont
    \setlength{\tabcolsep}{3pt}
    \renewcommand{\arraystretch}{.6}
    \renewcommand{\aboverulesep}{0.3ex}  
    \renewcommand{\belowrulesep}{0.5ex}  
    \begin{subtable}[h]{0.32\textwidth}
        \begin{tabularx}{\linewidth}{rXcc}
            \toprule
                   & Team             & macro          & micro                \\\midrule
                    1  & UMUTeam            &         81.95  &         82.00  \\
                       & SheffieldVeraAI               &         81.95  &         82.00  \\
                       & \quad:                  &         :      &         :      \\
                    5  & MELODI             &         77.89  &         78.00  \\
            \textbf{6} & \textbf{Hitachi}   & \textbf{77.66} & \textbf{76.00} \\
                    7  & FTD &         71.27  &         72.00  \\
            \bottomrule
        \end{tabularx}
        \caption{German (15 teams)}\label{tab:experiment-leaderboard-subtask1-de}
    \end{subtable}~~~
    \begin{subtable}[h]{0.32\textwidth}
        \begin{tabularx}{\linewidth}{rXcc}
            \toprule
                       & Team              & macro          & micro          \\\midrule
                    1  & MELODI            &         78.43  &         81.48  \\
                    2  & MLModeler5        &         61.63  &         62.96  \\
                       & \quad :           &         :      &         :      \\
                    6  & Unisa             &         58.62  &         61.11  \\
            \textbf{7} & \textbf{Hitachi}  & \textbf{55.29} & \textbf{59.26} \\
                    8  & UnedMediaBiasTeam &         52.36  &         57.41  \\
            \bottomrule
        \end{tabularx}
        \caption{English (22 teams)}\label{tab:experiment-leaderboard-subtask1-en}
    \end{subtable}~~~
    \begin{subtable}[h]{0.32\textwidth}
        \begin{tabularx}{\linewidth}{rXcc}
            \toprule
                       & Team             & macro          & micro          \\\midrule
                    1  & UMUTeam            &         83.55  &         88.00  \\
                    2  & QCRI           &         76.74  &         80.00  \\
            \textbf{3} & \textbf{Hitachi}   & \textbf{74.36} & \textbf{78.00} \\
                    4  & DSHacker           &         71.05  &         72.00  \\
                    5  & SheffieldVeraAI               &         68.16  &         74.00  \\
                    6  & FTD &         67.14  &         78.00  \\
            \bottomrule
        \end{tabularx}
        \caption{French (16 teams)}\label{tab:experiment-leaderboard-subtask1-fr}
    \end{subtable}\\
    \begin{subtable}[h]{0.32\textwidth}
        \begin{tabularx}{\linewidth}{rXcc}
            \toprule
                       & Team             & macro          & micro          \\\midrule
            \textbf{1} & \textbf{Hitachi}  & \textbf{76.83} & \textbf{85.25} \\
                    2  & QUST              &         76.68  &         83.61  \\
                    3  & DSHacker          &         72.04  &         83.61  \\
                       & SheffieldVeraAI              &         72.04  &         83.61  \\
                    5  & MELODI            &         58.67  &         75.41  \\
                    6  & UnedMediaBiasTeam &         58.41  &         62.30  \\
            \bottomrule
        \end{tabularx}
        \caption{Italian (16 teams)}\label{tab:experiment-leaderboard-subtask1-it}
    \end{subtable}~~~
    \begin{subtable}[h]{0.32\textwidth}
        \begin{tabularx}{\linewidth}{rXcc}
            \toprule
                       & Team               & macro          & micro          \\\midrule
                    1  & FTD &         78.55  &         93.62  \\
            \textbf{2} & \textbf{Hitachi}   & \textbf{77.92} & \textbf{87.23} \\
                    3  & SheffieldVeraAI               &         76.45  &         85.11  \\
                    4  & MELODI             &         70.86  &         85.11  \\
                    5  & UMUTeam            &         66.43  &         80.85  \\
                    6  & SinaaAI            &         66.35  &         80.85  \\
                \bottomrule
        \end{tabularx}
        \caption{Polish (16 teams)}\label{tab:experiment-leaderboard-subtask1-po}
    \end{subtable}~~~
    \begin{subtable}[h]{0.32\textwidth}
        \begin{tabularx}{\linewidth}{rXcc}
            \toprule
                       & Team               & macro          & micro          \\\midrule
            \textbf{1} & \textbf{Hitachi}   & \textbf{75.49} & \textbf{75.00} \\
                    2  & SheffieldVeraAI               &         72.87  &         72.22  \\
                    3  & FTD &         66.80  &         69.44  \\
                    4  & UMUTeam            &         64.54  &         68.06  \\
                    5  & MELODI             &         58.64  &         62.50  \\
                    6  & QCRI           &         56.66  &         65.28  \\
            \bottomrule
        \end{tabularx}
        \caption{Russian (16 teams)}\label{tab:experiment-leaderboard-subtask1-ru}
    \end{subtable}
    \caption{An excerpt from the official leaderboard for subtask 1 showing the rank, macro-F1 and micro-F1 on the single official run on the test split in each language}\label{tab:experiment-leaderboard-subtask1}
\end{table*}

\begin{table}[t]
    \centering
    \fontsize{8pt}{10pt}\selectfont
    \setlength{\tabcolsep}{4pt}
    \renewcommand{\arraystretch}{.6}
    \renewcommand{\aboverulesep}{0.3ex}  
    \renewcommand{\belowrulesep}{0.5ex}  
    \begin{tabular}{rlcc}
        \toprule
                   & Team             & macro F1       & micro F1       \\\midrule
        \textbf{1} & \textbf{Hitachi} & \textbf{72.93} & \textbf{76.79} \\
                2  & SheffieldVeraAI             &         72.13  &         77.88  \\
                3  & MELODI           &         68.35  &         76.08  \\
                4  & DSHacker         &         67.58  &         73.52  \\
                5  & UMUTeam          &         65.52  &         75.60  \\
                6  & MLModeler5       &         61.63  &         62.96  \\
        \bottomrule
    \end{tabular}
    \caption{An unofficial subtask 1 leaderboard sorted by the mean macro-F1 over six languages (from \cref{tab:experiment-leaderboard-subtask1}) }\label{tab:experiment-leaderboard-subtask1-all}
\end{table}

For each language-subtask pair, we picked four models from each group, resulting in 12 models for each language-subtask pair.
The four models were chosen whose macro-F1, micro-F1, ROC-AUC or mAP was the best in the development dataset for the target language-subtask pair.
This means that the same model can be chosen multiple times (e.g., a model which was the best in macro-F1 and micro-F1).
We did not remove the duplicates in that case --- such model will be sampled twice as much as a model which was the best only in a single metric.

Regarding these Stage I models and vanilla PLMs as an additional hyperparameter, we carried out Stage II random hyperparameter search on each language.
We sampled Stage I models three times more than PLMs, so that all groups (i.e., the four arrows entering ``en subtask 1'' in \cref{fig:training}) are sampled equally.
We trained 50 models for each language-subtask pair (50 models $\times$ 6 languages $\times$ 2 subtasks $=$ 600 models).

\begin{figure*}[t]
    \centering
    \includegraphics{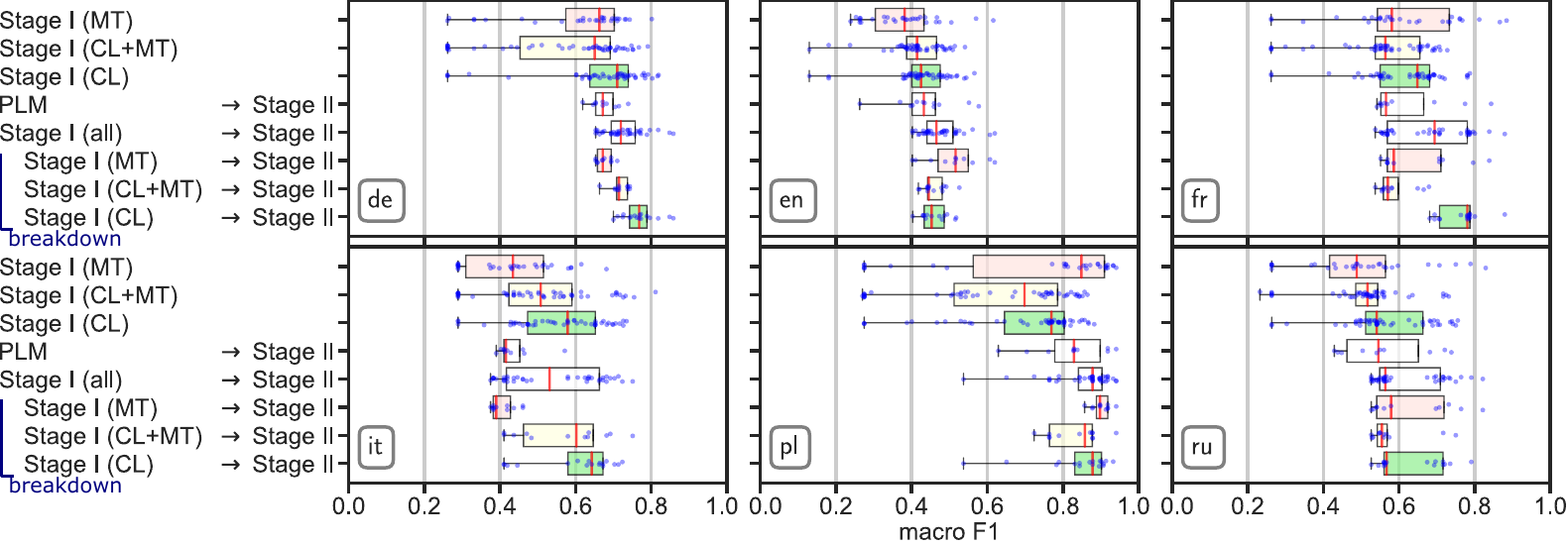}
    \caption{Comparison of subtask 1 macro-F1 (the development dataset) under different training paradigms (CL: cross-lingual/MT: multi-task)}\label{fig:experiment-subtask1}
\end{figure*}

\begin{figure}[t]
    \centering
    \includegraphics{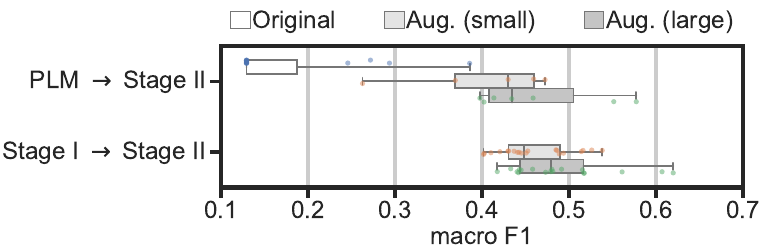}
    \caption{The effect of different training datasets in the development dataset for English subtask 1 }\label{fig:experiment-subtask1-en-dataset}
\end{figure}

\subsection{Ensembling}\label{sec:framework-ensemble}

Finally, we created an ensemble for each language-subtask pair from the results of hyperparameter search.
In a rare case, fine-tuning the model on the downstream task can degrade the performance.
Hence, we also considered the Stage I models for the ensemble.

We implemented multiple ensemble methods.
Due the scarcity of the development data, the results tend to be unstable.
Hence, we manually chose the best one for each language-subtask pair while monitoring multiple leave-one-out metrics on the development dataset.
This allowed us to choose models that are not only overfitting to a single metric.
The details of ensembling are described in \cref{sec:appendix-ensemble}.

\section{Results}\label{sec:experiments}

\subsection{Subtask 1: News Genre Categorization}\label{sec:experiment-subtask1}

Excerpts from the official leaderboards \cite{semeval2023task3} for subtask 1 are shown in \cref{tab:experiment-leaderboard-subtask1}.
We were the first place in Italian and Russian and within top threes in French and Polish.
In an unofficial ranking of mean macro-F1 of six languages, we were the first place (\cref{tab:experiment-leaderboard-subtask1-all}).

In \cref{fig:experiment-subtask1}, we show macro-F1 for the development dataset of all the models considered for the ensemble construction.
In all six languages, the models fine-tuned from cross-lingual and/or multi-lingual pretraining tend to perform better (i.e., have better median macro-F1) than the single language/task models trained from PLM (``PLM $\rightarrow$ Stage II'').
This shows that cross-lingual multi-task training was overall useful for genre categorization.
In most cases, fine-tuning Stage I models in Stage II yields better results than the vanilla Stage I models.

The breakdown of the performance based on how each model was pretrained in Stage I is also shown in the \cref{fig:experiment-subtask1}.
The results are mixed as to which Stage I pretraining paradigms were useful to the Stage II downstream performance.
In German, French and Italian, cross-lingual pretraining tends to be more beneficial than multi-task pretraining.
In English, Polish and Russian, multi-task pretraining tends to be more beneficial.
Interestingly, the combination of the both was never the best option in any language.

We analyzed the effect of incorporating external, balanced datasets for English subtask 1 (\cref{fig:experiment-subtask1-en-dataset}).
When directly fine-tuning PLM in Stage II, we see that models using either of external datasets tend to be considerably better than ones trained on the original training data.
For both conditions, we can see that Augmented (large) tend to perform better than Augmented (small).
This shows that obtaining a balanced dataset is important for news genre categorization.

\subsection{Subtask 2: Framing Detection}\label{sec:experiment-subtask2}

Excerpts from the official leaderboards \cite{semeval2023task3} for subtask 2 are shown in \cref{tab:experiment-leaderboard-subtask2}.
We were third to fifth places in all but Russian where we obtained the ninth place.

\begin{figure*}[t]
    \centering
    \centering
    \includegraphics{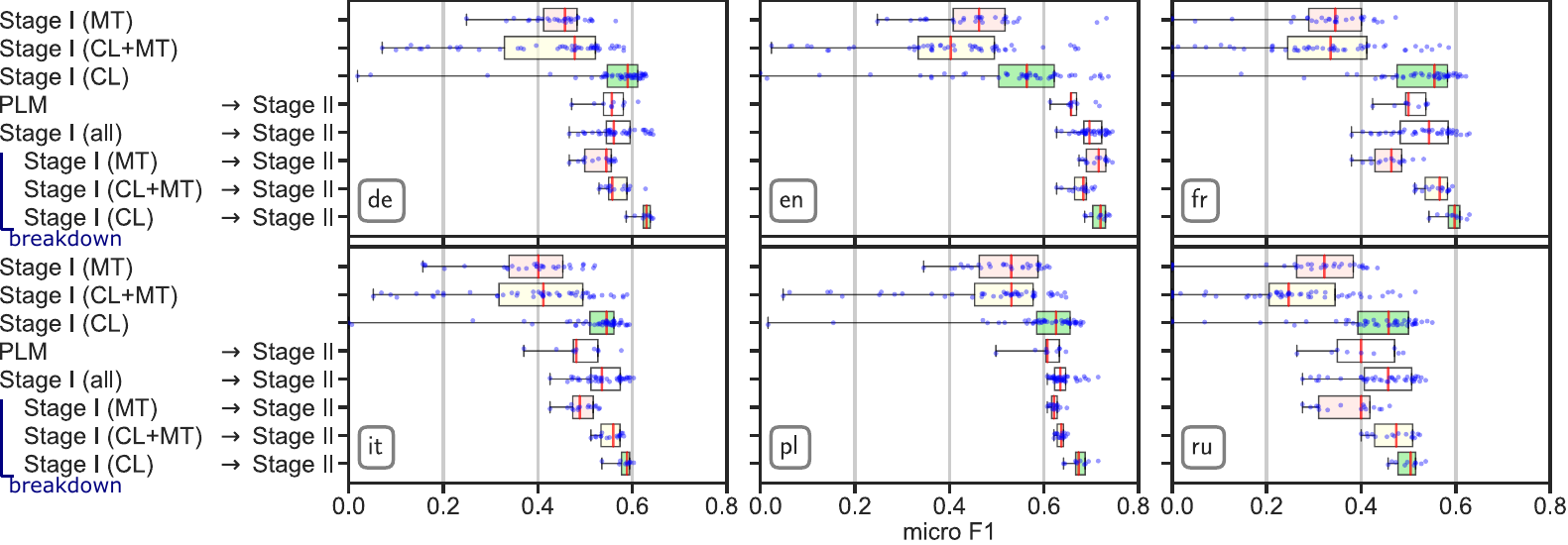}
    \caption{Comparison of subtask 2 micro-F1 (the develpment dataset) under different training paradigms (CL: cross-lingual/MT: multi-task)}\label{fig:experiment-subtask2}
\end{figure*}

In \cref{fig:experiment-subtask2}, we show micro-F1 for the development dataset of all the models considered for the ensemble construction.
As in subtask 1, the models fine-tuned from cross-lingual and/or multi-lingual pretraining tend to perform better than the single language/task models fine-tuned directly from PLM.
This suggests that cross-lingual multi-task training was also useful for framing detection.

In all languages in subtask 2, cross-lingual Stage II pretraining tends to result in better micro-F1 than multi-task models (\cref{fig:experiment-subtask2}).
We suspect that this lies in the difference in the linguistic nature of two subtasks;
Framing can be determined by lexical semantic to some extent, hence transfers well across different languages with multilingual transformers.
On the other hand, distinguishing the genre requires capturing language-specific pragmatics which may be the reason why it did not transfer between languages as effectively as subtask 2.

\begin{table*}[t]
    \centering
    \fontsize{8pt}{10pt}\selectfont
    \setlength{\tabcolsep}{3pt}
    \renewcommand{\arraystretch}{.6}
    \renewcommand{\aboverulesep}{0.3ex}  
    \renewcommand{\belowrulesep}{0.5ex}  
    \begin{subtable}[h]{0.32\textwidth}
        \begin{tabularx}{\linewidth}{rXcc}
            \toprule
                       & Team             & micro          & macro          \\\midrule
                    1  & MarsEclipse      &         71.12  &         66.05  \\
                    2  & QCRI         &         66.02  &         60.56  \\
                    3  & SheffieldVeraAI             &         65.25  &         60.14  \\
                    4  & TeamAmpa         &         63.22  &         57.27  \\
            \textbf{5} & \textbf{Hitachi} & \textbf{62.91} & \textbf{56.73} \\
                    6  & mCPTP         &         62.22  &         56.44  \\
            \bottomrule
        \end{tabularx}
        \caption{German (18 teams)}\label{tab:experiment-leaderboard-subtask2-de}
    \end{subtable}~~~
    \begin{subtable}[h]{0.32\textwidth}
        \begin{tabularx}{\linewidth}{rXcc}
            \toprule
                       & Team             & micro          & macro          \\\midrule
                    1  & SheffieldVeraAI             &         57.89  &         53.90  \\
                    2  & TeamAmpa         &         56.70  &         50.96  \\
                    3  & MarsEclipse      &         56.23  &         49.05  \\
            \textbf{4} & \textbf{Hitachi} & \textbf{54.26} & \textbf{47.16} \\
                    5  & mCPTP         &         53.53  &         48.17  \\
                    6  & QUST             &         51.31  &         46.21  \\
            \bottomrule
        \end{tabularx}
        \caption{English (22 teams)}\label{tab:experiment-leaderboard-subtask2-en}
    \end{subtable}~~~
    \begin{subtable}[h]{0.32\textwidth}
        \begin{tabularx}{\linewidth}{rXcc}
            \toprule
                       & Team             & micro          & macro          \\\midrule
                    1  & MarsEclipse      &         55.28  &         53.68  \\
                    2  & BERTastic        &         53.69  &         52.02  \\
                    3  & SheffieldVeraAI             &         53.42  &         52.03  \\
            \textbf{4} & \textbf{Hitachi} & \textbf{51.41} & \textbf{48.83} \\
                    5  & TeamAmpa         &         50.56  &         47.89  \\
                    6  & TheSyllogist     &         48.57  &         46.16  \\
            \bottomrule
        \end{tabularx}
        \caption{French (18 teams)}\label{tab:experiment-leaderboard-subtask2-fr}
    \end{subtable}\\
    \begin{subtable}[h]{0.32\textwidth}
        \begin{tabularx}{\linewidth}{rXcc}
            \toprule
                       & Team             & micro          & macro          \\\midrule
                    1  & MarsEclipse      &         61.73  &         54.46  \\
                    2  & QCRI         &         59.91  &         47.95  \\
            \textbf{3} & \textbf{Hitachi} & \textbf{59.77} & \textbf{51.51} \\
                    4  & TeamAmpa         &         59.67  &         48.27  \\
                    5  & mCPTP         &         58.41  &         46.88  \\
                    6  & UMUTeam          &         57.63  &         44.67  \\
            \bottomrule
        \end{tabularx}
        \caption{Italian (18 teams)}\label{tab:experiment-leaderboard-subtask2-it}
    \end{subtable}~~~
    \begin{subtable}[h]{0.32\textwidth}
        \begin{tabularx}{\linewidth}{rXcc}
            \toprule
                       & Team             & micro          & macro          \\\midrule
                    1  & MarsEclipse      &         67.31  &         63.84  \\
                    2  & SheffieldVeraAI             &         64.52  &         60.27  \\
                    3  & QCRI         &         64.19  &         59.87  \\
                    4  & UMUTeam          &         64.18  &         59.31  \\
            \textbf{5} & \textbf{Hitachi} & \textbf{63.40} & \textbf{58.40} \\
                    6  & SATLab           &         62.02  &         56.99  \\
                \bottomrule
        \end{tabularx}
        \caption{Polish (18 teams)}\label{tab:experiment-leaderboard-subtask2-po}
    \end{subtable}~~~
    \begin{subtable}[h]{0.32\textwidth}
        \begin{tabularx}{\linewidth}{rXcc}
            \toprule
                       & Team             & micro          & macro          \\\midrule
                    1  & MarsEclipse      &         44.98  &         30.33  \\
                    2  & SheffieldVeraAI             &         44.14  &         35.59  \\
                       & \quad:           &         :      &         :      \\
                    8  & UMUTeam          &         38.49  &         28.84  \\
            \textbf{9} & \textbf{Hitachi} & \textbf{37.00} & \textbf{32.59} \\
                    10  & Riga             &         31.51  &         22.19 \\
            \bottomrule
        \end{tabularx}
        \caption{Russian (17 teams)}\label{tab:experiment-leaderboard-subtask2-ru}
    \end{subtable}
    \caption{An excerpt from the official leaderboard for subtask 2 showing the rank, micro-F1 and macro-F1 on the single official run on the test split in each language}\label{tab:experiment-leaderboard-subtask2}
\end{table*}

\begin{table}[t]
    \centering
    \fontsize{8pt}{10pt}\selectfont
    \setlength{\tabcolsep}{4pt}
    \renewcommand{\arraystretch}{.6}
    \renewcommand{\aboverulesep}{0.3ex}  
    \renewcommand{\belowrulesep}{0.5ex}  
    \begin{tabular}{rlcc}
        \toprule
                   & Team             & micro F1       & macro F1       \\\midrule
                1  & MarsEclipse      &         59.44  &         52.90  \\
                2  & SheffieldVeraAI             &         57.05  &         51.85  \\
                3  & QCRI         &         55.47  &         48.29  \\
                4  & TeamAmpa         &         55.41  &         48.22  \\
        \textbf{5} & \textbf{Hitachi} & \textbf{54.79} & \textbf{49.20} \\
                6  & mCPTP         &         53.60  &         47.74  \\
        \bottomrule
    \end{tabular}
    \caption{An unofficial subtask 2 leaderboard sorted by the mean micro-F1 in six languages (from \cref{tab:experiment-leaderboard-subtask1})}\label{tab:experiment-leaderboard-subtask2-all}
\end{table}

\section{Related Work}\label{sec:relatedwork}

There exist several studies on multilingual and/or multi-task learning in the context of media analysis.
\citet{uyangodage2021can} applied cross-lingual training to multilingual false information detection and showed that cross-lingual training results in on par performance to monolingual training.
\citet{alam2021fighting,alam2020fighting} annotated Tweets in multiple languages with multiple choice questions regarding COVID-19 disinformation.
They fine-tuned mBERT model \cite{devlin_bert_2019} in a cross-lingual setup and in a multi-task setup, and found that the results are mixed in terms of their benefits.
The SemEval 2023 Task 3 dataset \cite{semeval2023task3} exhibits significantly different properties to the datasets previous works used and we found that cross-lingual and/or multi-task learning help in this dataset.
We also carried out more thorough experiments than the previous works, which might have been the key to the improvement.

There also exist different approaches for multilingual and low-resource settings in broader domains.
\citet{reimers2020making} proposed a method for training multilingual sentence embeddings and demonstrated its effectiveness in 50+ languages.
In a low-resource setup, \citet{heinisch2022data} used a data augmentation approach that retrieves additional data from related datasets by automatically labeling them.
We only explored BERT-based approach in this work, but we wish to explore other approaches for cross-lingual multi-task learning in the future.

\section{Conclusion}\label{sec:conclusion}

In our participation to SemEval-2023 Task 3, we investigated different strategies for multilingual genre and framing detection.
Through the extensive experiments, we found that collecting an external balanced dataset can help genre categorization.
We also find that cross-lingual and multi-task training can help both genre and framing detection and found that cross-lingual training is more beneficial for framing detection.
We constructed ensemble models from the results and achieved the highest macro-F1 in Italian and Russian genre detection.

For future work, we will investigate the effect of cross-lingual multi-task training on zero-shot language transfer (Greek, Spanish and Georgian subtasks that we did not participate), as well as the effect on and benefit from training models on persuasion techniques detection (subtask 3).

\section*{Acknowledgements}

We used computational resource of AI Bridging Cloud Infrastructure (ABCI) provided by the National Institute of Advanced Industrial Science and Technology (AIST) for the experiments.

We would like to thank Gaku Morio and Yuichi Sasazawa for their valuable feedbacks.
We would also like to thank Dr. Masaaki Shimizu for helping us with the computational resources.

\bibliography{custom}
\bibliographystyle{acl_natbib}

\appendix

\section{Appendix}\label{sec:appendix}

\subsection{Data Analysis}\label{sec:appendix-task}

\begin{table}[b]
    \centering
    \fontsize{8pt}{10pt}\selectfont
    \setlength{\tabcolsep}{4pt}
    \renewcommand{\arraystretch}{.6}
    \renewcommand{\aboverulesep}{0.3ex}  
    \renewcommand{\belowrulesep}{0.5ex}  
    \begin{tabular}{lrrr}
        \toprule
        Language     & Subtask 1 & Subtask 2 & Overlap \\ \midrule
        Germany (de) & 132 & 132 &  97 \\
        English (en) & 433 & 433 & 433 \\
        French (fr)  & 157 & 158 & 119 \\
        Italian (it) & 226 & 227 & 170 \\
        Polish (pl)  & 144 & 145 & 106 \\
        Russiun (ru) & 142 & 143 & 107 \\
        \bottomrule
    \end{tabular}
    \caption{Numbers of articles in subtask 1 and 2 training data, and their overlaps}\label{tab:document-overlap}
\end{table}

Numbers of articles in each subtask and their overlaps are summarized in \cref{tab:document-overlap}.
We can see that the numbers of articles are limited for the relative large label space and there exist considerable overlaps of articles between subtask 1 and 2.

Readers should refer \cite{semeval2023task3} for the details on the datasets.

\subsection{Details of Hyperparameter Search}\label{sec:appendix-hyperparameters}

\begin{table}[t]
    \centering
    \fontsize{8pt}{10pt}\selectfont
    \setlength{\tabcolsep}{4pt}
    \renewcommand{\arraystretch}{.8}
    \begin{tabularx}{\linewidth}{llX}
        \toprule
        \multicolumn{2}{l}{Hyperparameter} & Values\\\midrule
        \multicolumn{2}{l}{Base model} & XLM-RoBERTa large, RemBERT\\
        \multicolumn{2}{l}{Dataset} & All, All but English \\
        \multicolumn{2}{l}{Classwise training} & No \\
        \multicolumn{2}{l}{Max steps} & 100, 200, 300 \\
        \multicolumn{2}{l}{Learning rate} & 30, 20, 15, 10, 8, 5\quad($\times 10^{\text{-}6}$)\\
        \multicolumn{2}{l}{Batch size} & 32, 64 \\
        \multicolumn{2}{l}{Weight decay} &  0.02, 0.01, 0.001 \\
        \multicolumn{2}{l}{Loss scaling} & Yes, No \\
        \multicolumn{2}{l}{Loss scale threshold} & N/A, 5, 10,000 \\
        \multicolumn{2}{l}{Gradient clipping} &  1.0 \\
        \multicolumn{2}{l}{Warmup ratio} &  0.2 \\
        \bottomrule
    \end{tabularx}
    \caption{Hyperparamter search space for Stage I cross-lingual multi-task training}\label{tab:hyperparameters-all}
\end{table}

\begin{table}[t]
    \centering
    \fontsize{8pt}{10pt}\selectfont
    \setlength{\tabcolsep}{4pt}
    \renewcommand{\arraystretch}{.8}
    \begin{tabularx}{\linewidth}{llX}
        \toprule
        \multicolumn{2}{l}{Hyperparameter} & Values\\\midrule
        \multicolumn{2}{l}{Base model} & XLM-RoBERTa large, RemBERT\\
        \multicolumn{2}{l}{Dataset} & All, All but English \\
        \multicolumn{2}{l}{Max steps} & 100, 200, 300 \\
        \multicolumn{2}{l}{Learning rate} & 30, 20, 15, 10, 8, 5\quad($\times 10^{\text{-}6}$)\\
        \multicolumn{2}{l}{Batch size} & 32, 64 \\
        \multicolumn{2}{l}{Weight decay} &  0.02, 0.01, 0.001 \\
        \multicolumn{2}{l}{Loss scaling} & Yes, No \\
        \multicolumn{2}{l}{Gradient clipping} &  1.0 \\
        \multicolumn{2}{l}{Warmup ratio} &  0.2 \\
        \bottomrule
    \end{tabularx}
    \caption{Hyperparamter search space for Stage I cross-lingual training subtask 1}\label{tab:hyperparameters-subtask1}
\end{table}

\begin{table}[!t]
    \centering
    \fontsize{8pt}{10pt}\selectfont
    \setlength{\tabcolsep}{4pt}
    \renewcommand{\arraystretch}{.8}
    \begin{tabularx}{\linewidth}{llX}
        \toprule
        \multicolumn{2}{l}{Hyperparameter} & Values\\\midrule
        \multicolumn{2}{l}{Base model} & XLM-RoBERTa large, RemBERT\\
        \multicolumn{2}{l}{Dataset} & All, All but English \\
        \multicolumn{2}{l}{Classwise training} & No \\
        \multicolumn{2}{l}{Max steps} & 100, 200, 300 \\
        \multicolumn{2}{l}{Learning rate} & 30, 20, 15, 10, 8, 5\quad($\times 10^{\text{-}6}$)\\
        \multicolumn{2}{l}{Batch size} & 32, 64 \\
        \multicolumn{2}{l}{Weight decay} &  0.02, 0.01, 0.001 \\
        \multicolumn{2}{l}{Loss scale threshold} & N/A, 5, 10,000 \\
        \multicolumn{2}{l}{Gradient clipping} &  1.0 \\
        \multicolumn{2}{l}{Warmup ratio} &  0.2 \\
        \bottomrule
    \end{tabularx}
    \caption{Hyperparamter search space for Stage I cross-lingual training of subtask 2}\label{tab:hyperparameters-subtask2}
\end{table}

\begin{table}[t]
    \centering
    \fontsize{8pt}{10pt}\selectfont
    \setlength{\tabcolsep}{4pt}
    \renewcommand{\arraystretch}{.8}
    \begin{tabularx}{\linewidth}{llX}
        \toprule
        \multicolumn{2}{l}{Hyperparameter} & Values\\\midrule
        \multicolumn{2}{l}{Base model} & RoBERTa large \\
        \multicolumn{2}{l}{Dataset} & Aug. (large) + official subtask 2 dataset, Aug. (small) + official subtask 2 dataset \\
        \multicolumn{2}{l}{Max steps} & 80, 120, 160, 200 \\
        \multicolumn{2}{l}{Learning rate} &  80, 6, 5, 4, 2\quad($\times 10^{\text{-}6}$) \\
        \multicolumn{2}{l}{Batch size} & 32, 64 \\
        \multicolumn{2}{l}{Weight decay} &  0.02, 0.01, 0.001 \\
        \multicolumn{2}{l}{Loss scaling} & Yes, No \\
        \multicolumn{2}{l}{Loss scale threshold} & N/A, 5, 10,000 \\
        \multicolumn{2}{l}{Gradient clipping} &  1.0 \\
        \multicolumn{2}{l}{Warmup ratio} &  0.2 \\
        \bottomrule
    \end{tabularx}
    \caption{Hyperparamter search space for Stage I multi-task training of English subtask 1 and 2}\label{tab:hyperparameters-en}
\end{table}

\begin{table}[t]
    \centering
    \fontsize{8pt}{10pt}\selectfont
    \setlength{\tabcolsep}{4pt}
    \renewcommand{\arraystretch}{.8}
    \begin{tabularx}{\linewidth}{llX}
        \toprule
        \multicolumn{2}{l}{Hyperparameter} & Values\\\midrule
        \multicolumn{2}{l}{Base model} & XLM-RoBERTa large, RemBERT \\
        \multicolumn{2}{l}{Dataset} & Official subtask 1 and 2 datasets in each language \\
        \multicolumn{2}{l}{Max steps} & 80, 120, 160, 200 \\
        \multicolumn{2}{l}{Learning rate} &  15, 12, 10, 8, 6, 4\quad($\times 10^{\text{-}6}$) \\
        \multicolumn{2}{l}{Batch size} & 16, 32 \\
        \multicolumn{2}{l}{Weight decay} &  0.02, 0.01, 0.001 \\
        \multicolumn{2}{l}{Loss scaling} & Yes, No \\
        \multicolumn{2}{l}{Loss scale threshold} & N/A, 5, 10,000 \\
        \multicolumn{2}{l}{Gradient clipping} &  1.0 \\
        \multicolumn{2}{l}{Warmup ratio} &  0.2 \\
        \bottomrule
    \end{tabularx}
    \caption{Hyperparamter search space for Stage I multi-task training of a single language (German, French, Italian, Polish or Russian)}\label{tab:hyperparameters-others}
\end{table}

As described in \cref{sec:framework}, we carried out the extensive experiments as a series of hyperparameter searches.
In this section, we will list and describe all the hyperparamter search spaces of Stage I and II training.

The hyperparameter search spaces of Stage I training are listed as in the following:
\begin{itemize}
    \item Cross-lingual multi-task: \cref{tab:hyperparameters-all}
    \item Cross-lingual \cref{tab:hyperparameters-subtask1} and \ref{tab:hyperparameters-subtask2}
    \item Multi-task: \cref{tab:hyperparameters-en} and \ref{tab:hyperparameters-others}
\end{itemize}
The hyperparameter search spaces of Stage II training are listed in the following:
\begin{itemize}
    \item Subtask 1 in English: \cref{tab:hyperparameters-en-subtask1}
    \item Subtask 1 in all other languages: \cref{tab:hyperparameters-others-subtask1}
    \item Subtask 2 in English: \cref{tab:hyperparameters-en-subtask2}
    \item Subtask 2 in all other languages: \cref{tab:hyperparameters-others-subtask2}
\end{itemize}

We introduced a loss weighting technique and introduced it as an additional hyperparameters.
Since label distributions are highly skewed in subtask 1, we weight losses for each label $\mathcal{L}_l$ ($l \in \{satire, opinion, reporting\}$) by $w_l$ (i.e., $\mathcal{L}'_l = w_l \cdot \mathcal{L}_l$) such that they are inversely proportional to the count of each label $c_l$ (i.e., $w_l \propto 1 / c_l$) while adding up to 1 (i.e., $\sum\nolimits_l w_l = 1$).
\begin{equation}
    w_l = \frac{\text{hmean}(c_{satire} + c_{opinion} + c_{reporting})}{c_l},
\end{equation}
where $\text{hmean}$ is harmonic mean.

For subtask 2, we carried out the classification in both a single multi-label classification and multiple, separate binary classifications.
We have also regarded the choice of the classification method as a hyperparamter and incorporated this into the random search in Stage II (``classwise'').

\begin{table}[t]
    \centering
    \fontsize{8pt}{10pt}\selectfont
    \setlength{\tabcolsep}{4pt}
    \renewcommand{\arraystretch}{.8}
    \begin{tabularx}{\linewidth}{llX}
        \toprule
        \multicolumn{2}{l}{Hyperparameter} & Values\\\midrule
        \multicolumn{2}{l}{Base model} & RoBERTa large, Stage I models \\
        \multicolumn{2}{l}{Dataset} & Aug. (small), Aug. (large) \\
        \multicolumn{2}{l}{Max steps} & \\
        & if PLM & 100, 150, 200 \\
        & if Stage I & 30, 50, 80, 100, 150 \\
        \multicolumn{2}{l}{Learning rate} & \\
        & if PLM & 20, 15, 10, 8, 6, 4, 3, 2\quad($\times 10^{\text{-}6}$)\\
        & if Stage I & 10, 8, 6, 5, 4, 2\quad($\times 10^{\text{-}6}$)\\
        \multicolumn{2}{l}{Batch size} & 16, 32 \\
        \multicolumn{2}{l}{Weight decay} &  0.02, 0.01, 0.001 \\
        \multicolumn{2}{l}{Gradient clipping} &  1.0 \\
        \multicolumn{2}{l}{Warmup ratio} &  0.2 \\
        \bottomrule
    \end{tabularx}
    \caption{Hyperparamter search space for Stage II training of English subtask 1}\label{tab:hyperparameters-en-subtask1}
\end{table}

\begin{table}[t]
    \centering
    \fontsize{8pt}{10pt}\selectfont
    \setlength{\tabcolsep}{4pt}
    \renewcommand{\arraystretch}{.8}
    \begin{tabularx}{\linewidth}{llX}
        \toprule
        \multicolumn{2}{l}{Hyperparameter} & Values\\\midrule
        \multicolumn{2}{l}{Base model} & XLM-RoBERTa large, RemBERT, Stage 1 models \\
        \multicolumn{2}{l}{Dataset} & Official dataset for each language \\
        \multicolumn{2}{l}{Max steps} & \\
        & if PLM & 160, 200, 240 \\
        & if Stage I & 30, 50, 80, 100, 150 \\
        \multicolumn{2}{l}{Learning rate} & \\
        & if PLM  & 15, 12, 10, 8, 5\quad($\times 10^{\text{-}6}$)\\
        & if Stage I & 10, 8, 6, 5, 4, 2\quad($\times 10^{\text{-}6}$)\\
        \multicolumn{2}{l}{Batch size} & 16, 32 \\
        \multicolumn{2}{l}{Weight decay} &  0.02, 0.01, 0.001 \\
        \multicolumn{2}{l}{Loss scaling)} & Yes, No\\
        \multicolumn{2}{l}{Gradient clipping} &  1.0 \\
        \multicolumn{2}{l}{Warmup ratio} &  0.2 \\
        \bottomrule
    \end{tabularx}
    \caption{Hyperparamter search space for Stage II training of German, French, Italian, Polish and Russian subtask 1}\label{tab:hyperparameters-others-subtask1}
\end{table}

\begin{table}[t]
    \centering
    \fontsize{8pt}{10pt}\selectfont
    \setlength{\tabcolsep}{4pt}
    \renewcommand{\arraystretch}{.8}
    \begin{tabularx}{\linewidth}{llX}
        \toprule
        \multicolumn{2}{l}{Hyperparameter} & Values\\\midrule
        \multicolumn{2}{l}{Base model} & RoBERTa large, Stage I models \\
        \multicolumn{2}{l}{Dataset} & Official dataset \\
        \multicolumn{2}{l}{Classwise training} & Yes, No \\
        \multicolumn{2}{l}{Max steps} & \\
        & if PLM & 100, 150, 200 \\
        & if Stage I w/ classwise & 80, 100, 120, 140 \\
        & if Stage I w/o classwise & 80, 100, 120, 140, 180 \\
        \multicolumn{2}{l}{Learning rate} & \\
        & if for PLM  & 20, 15, 12, 10, 8, 6\quad($\times 10^{\text{-}6}$)\\
        & if Stage I & 15, 12, 10, 8, 6, 4, 1\quad($\times 10^{\text{-}6}$)\\
        \multicolumn{2}{l}{Batch size} & 16, 32 \\
        \multicolumn{2}{l}{Weight decay} &  0.02, 0.01, 0.001 \\
        \multicolumn{2}{l}{Loss scale threshold} & N/A, 5, 10,000 \\
        \multicolumn{2}{l}{Gradient clipping} &  1.0 \\
        \multicolumn{2}{l}{Warmup ratio} &  0.2 \\
        \bottomrule
    \end{tabularx}
    \caption{Hyperparamter search space for Stage II training of English subtask 2}\label{tab:hyperparameters-en-subtask2}
\end{table}

\begin{table}[t]
    \centering
    \fontsize{8pt}{10pt}\selectfont
    \setlength{\tabcolsep}{4pt}
    \renewcommand{\arraystretch}{.8}
    \begin{tabularx}{\linewidth}{llX}
        \toprule
        \multicolumn{2}{l}{Hyperparameter} & Values\\\midrule
        \multicolumn{2}{l}{Base model} & XLM-RoBERTa large, RemBERT, Stage I models \\
        \multicolumn{2}{l}{Dataset} & Official dataset \\
        \multicolumn{2}{l}{Classwise training} & Yes, No \\
        \multicolumn{2}{l}{Max steps} & \\
        & if PLM & 160, 200, 240, 280 \\
        & if Stage I w/ classwise & 80, 100, 120, 140 \\
        & if Stage I w/o classwise & 80, 100, 120, 140, 180 \\
        \multicolumn{2}{l}{Learning rate} & \\
        & if for PLM  & 15, 12, 10, 8, 6, 4\quad($\times 10^{\text{-}6}$)\\
        & if Stage I & 15, 12, 10, 8, 6, 4, 1\quad($\times 10^{\text{-}6}$)\\
        \multicolumn{2}{l}{Batch size} & 16, 32 \\
        \multicolumn{2}{l}{Weight decay} &  0.02, 0.01, 0.001 \\
        \multicolumn{2}{l}{Loss scale threshold} & N/A, 5, 10,000 \\
        \multicolumn{2}{l}{Gradient clipping} &  1.0 \\
        \multicolumn{2}{l}{Warmup ratio} &  0.2 \\
        \bottomrule
    \end{tabularx}
    \caption{Hyperparamter search space for Stage II training of German, French, Italian, Polish and Russian subtask 2}\label{tab:hyperparameters-others-subtask2}
\end{table}

\subsection{Details of Ensemble Construction}\label{sec:appendix-ensemble}

We created an ensemble for each language-subtask pair from the results of hyperparameter search.
As outlined in \cref{sec:framework-ensemble}, we implemented multiple ensemble methods and manually chose the best one for each language-subtask pair.
Here, we show the details of ensemble construction and selection on each subtask.

\subsubsection{Ensembles for Subtask 1}\label{sec:appendix-ensemble-subtask1}

For subtask 1, we implemented three ensemble methods:
\begin{description}
    \item[Top one] We choose the best model with the best macro-F1 in the development dataset.
    \item[Top 3 average] We picked three models based on the macro-F1 score in the development dataset. We take an average of the output probabilities (i.e., scores after softmax).
    \item[Bootstrap bagging] We greedily add models to average ensemble with replacement until the score no longer improves or the ensemble size reaches five. We use the minimum F1 score of all the classes. This idea of trying to improve the worst-class performance was inspired by distributionally robust optimization.
\end{description}

We also considered different pools of candidate models to construct the ensembles from.
Specifically, we either created an ensemble from \begin{enumerate*}[label=(\roman*)]
    \item all the models from Stage I and II, and
    \item the models only from Stage II (i.e., only models that are fine-tuned on the target language)
\end{enumerate*}.

\begin{table}[!t]
    \centering
    \begin{threeparttable}
        \centering
        \fontsize{8pt}{10pt}\selectfont
        \setlength{\tabcolsep}{4pt}
        \renewcommand{\arraystretch}{.8}
        \begin{tabular}{lllll}
            \toprule
             & \multicolumn{2}{c}{Ensemble} & \multicolumn{2}{c}{Postprocess} \\\cmidrule(lr){2-3}\cmidrule(lr){4-5}
            Language & Method & Candidates & Reweighting & Relabeling \\\midrule
            English & Top 3 & Stage I \& II & 1.5 & Yes \\
            French & Top 3 & Stage II & 1.5 & N/A \\
            German & Top 3 & Stage II & 1.5 & N/A \\
            Italian & Top 3 & Stage I \& II & 20.0 & N/A \\
            Polish & Bagging & Stage II & 1.5 & N/A \\
            Russian & Top 3 & Stage I \& II & 1.0 & N/A \\
            \bottomrule
        \end{tabular}
        \makeatletter\def\TPT@hsize{}\makeatletter
        \begin{tablenotes}[para,flushleft]
            \raggedright
            \fontsize{7pt}{7pt}\selectfont
            See \cref{sec:appendix-ensemble-subtask1} for the description of each column.
        \end{tablenotes}
    \end{threeparttable}
    \caption{The selected ensemble method for each language in subtask 1}\label{tab:ensemble-subtask1}
\end{table}

\begin{table*}[!t]
    \centering
    \begin{threeparttable}
        \centering
        \fontsize{8pt}{10pt}\selectfont
        \setlength{\tabcolsep}{4pt}
        \renewcommand{\arraystretch}{.8}
        \begin{tabularx}{\textwidth}{lXXXXXX}
            \toprule
             & \multicolumn{6}{c}{Language} \\\cmidrule(lr){2-7}
            Label & English & French & German & Italian & Polish & Russian \\\midrule
            Capacity and resources & Top3 (AP)* & Top3 (ROC) & Top1 (F1)* & Top5 (ROC)* & Top3 (ROC) & Top3 (F1)* \\
            Crime and punishment & Bagging* & Top5 (F1) & Bagging & Top3 (F1) & Top3 (ROC)* & Top5 (F1)* \\
            Cultural identity & Top5 (ROC)* & Top1 (F1)* & Top5 (F1) & Top5 (F1)* & Top5 (F1)* & Top3 (AP)* \\
            Economic & Bagging* & Top1 (F1) & Top5 (ROC) & Top5 (F1)* & Top3 (AP)* & Top3 (F1) \\
            External regulation and reputation & Top3 (F1)* & Top5 (F1)* & Top1 (F1) & Top3 (AP) & Top3 (AP) & Top3 (ROC)* \\
            Fairness and equality & Top3 (AP) & Bagging* & Top3 (AP) & Top3 (AP)* & Top3 (F1) & Top5 (F1) \\
            Health and safety & Top5 (ROC) & Top5 (ROC)* & Top3 (F1)* & Top5 (ROC)* & Top3 (ROC) & Top5 (F1) \\
            Legality, constitutionality and jurisprudence & Bagging & Top3 (ROC)* & Top3 (F1) & Top5 (F1)* & Top5 (F1)* & Top5 (F1)* \\
            Morality & Top5 (AP)* & Top3 (AP)* & Top3 (F1)* & Top5 (F1) & Top1 (F1) & Top3 (AP)* \\
            Policy prescription and evaluation & Stacking* & Top3 (AP)* & Top5 (ROC)* & Bagging* & Top3 (ROC)* & Top3 (AP)* \\
            Political & Top5 (F1)* & Top3 (AP) & Top3 (AP) & Top3 (F1) & Top3 (ROC)* & Top3 (AP) \\
            Public opinion & Top3 (F1)* & Top5 (F1)* & Top1 (F1)* & Top3 (F1) & Top5 (F1) & Top3 (AP) \\
            Quality of life & Top3 (AP)* & Top1 (F1) & Top5 (ROC)* & Top3 (F1) & Top3 (ROC) & Top3 (ROC)* \\
            Security and defense & Bagging* & Top3 (ROC) & Top3 (AP) & Top5 (F1) & Top5 (F1)* & Top5 (ROC) \\
            \bottomrule
        \end{tabularx}
        \makeatletter\def\TPT@hsize{}\makeatletter
        \begin{tablenotes}[para,flushleft]
            \raggedright
            \fontsize{7pt}{7pt}\selectfont
            * Chooses models from both Stage I and II. Otherwise, models are chosen only from Stage II. See \cref{sec:appendix-ensemble-subtask2} for the details about the ensemble methods.
        \end{tablenotes}
    \end{threeparttable}
    \caption{The selected ensemble method for each language in subtask 2}\label{tab:ensemble-subtask2}
\end{table*}

Due to the highly imbalanced label distribution, especially the lack of satire pieces in the training dataset, we introduced two postprocessing methods at the ensembling stage:
\begin{description}
    \item[Probability reweighting] We multiply the probability for satire label by a value, followed by renormalization of the probabilities by their sum.
    \item[Heuristics-based relabeling] The model tend to output the opinion label more than other labels in English, even though we balanced the training dataset (\cref{sec:dataset}). Hence,  we converted a portion of predicted opinion labels to other labels with heuristics. First, we selected  documents whose words consist of more than 0.8\% CARDINAL named entities\footnote{We used spaCy (\url{https://spacy.io/}) for named entity recognition.}. This is based on the intuition that both satire and reporting pieces tend to utilize numbers to be more persuasive. We then relabeled the selected documents with satire label if it contains ``!'' or ``?''. Otherwise, they were relabeled with reporting label.
\end{description}

The model pools and the postprocessing methods were also considered as an option and we compared all the combinations of the ensemble methods, the model pools and the postprocessing methods.

Due the scarcity of the development data, the results tend to be unstable.
Hence, we manually chose the best ensemble type for each language with following criteria while monitoring leave-one-out metrics on the development dataset.

\begin{itemize}
    \item We try to choose model with a good class score balance (i.e., good macro-F1) and good general classification abilities (good mAP and ROC-AUC).
    \item Unless the difference is unbearably large, we tried to avoid top one model as it can be unstable.
\end{itemize}

After choosing the best ensemble method, model pool and postprocessing method for each language, we recreated the ensemble using the whole development dataset.
The selected ensemble method for each language is shown in \cref{tab:ensemble-subtask1}.

\subsubsection{Ensembles for Subtask 2}\label{sec:appendix-ensemble-subtask2}

For subtask 2, we implemented nine ensemble methods:
\begin{description}
    \item[Top one] We choose the best model with the best macro-F1 in the development dataset.
    \item[Top $n$ average] We picked $n$ models based on the score in the development dataset. We take average of the output probabilities (i.e., score after the sigmoid). We adopted ranking by \begin{enumerate*}[label=(\arabic*)]
      \item F1 score with $n = 3$,
      \item average precision score with $n = 3$,
      \item ROC-AUC score with $n = 3$,
      \item F1 score with $n = 5$,
      \item average precision score with $n = 5$, and
      \item ROC-AUC score with $n = 5$
    \end{enumerate*}.
    \item[Bootstrap bagging] Same as subtask 1 but we optimize F1 score.
    \item[Stacking ensemble] We fit lasso regression classifier on the development dataset ($C = 1.0$), regarding probability from each model as a feature.
\end{description}
As in subtask 1, we considered different pools of candidate models to construct the ensembles from (\cref{sec:appendix-ensemble-subtask1}).

We manually chose the best ensemble method and model pool for each language-\emph{label} pair in the same way as subtask 1.
The selected ensemble method for each language-label pair is shown in \cref{tab:ensemble-subtask2}.

\end{document}